\documentclass[journal]{IEEEtran}

\usepackage{times}
\usepackage{epsfig}
\usepackage{graphicx}
\usepackage{amsmath}
\usepackage{amssymb}
\usepackage{verbatim}
\usepackage{subcaption}
\usepackage{float}
\usepackage{footmisc}
\usepackage{paralist}
\let\itemize\compactitem
\let\enditemize\endcompactitem
\let\enumerate\compactenum
\let\endenumerate\endcompactenum

\usepackage[pagebackref=true,breaklinks=true,colorlinks,bookmarks=false]{hyperref}

\newcommand\blfootnote[1]{
\begingroup
\renewcommand\thefootnote{}\footnote{#1}%
\addtocounter{footnote}{-1}%
\endgroup
}

\IEEEoverridecommandlockouts
\begin{document}

%%%%%%%%% TITLE
% \title{A Scientific ChildGAN With More Disentanglement}
\title{Heredity-aware Child Face Image Generation with Latent Space Disentanglement}

%\author{\IEEEauthorblockN{Michael Shell}
%\IEEEauthorblockA{School of Electrical and\\Computer Engineering\\
%Georgia Institute of Technology\\
%Atlanta, Georgia 30332--0250\\
%Email: http://www.michaelshell.org/contact.html}
%\and
%\IEEEauthorblockN{Homer Simpson}
%\IEEEauthorblockA{Twentieth Century Fox\\
%Email: homer@thesimpsons.com}}


\author{Xiao~Cui, Wengang~Zhou, Yang~Hu, Weilun~Wang
        and~Houqiang~Li,~\IEEEmembership{Fellow,~IEEE}% <-this % stops a space
\thanks{Xiao~Cui, Wengang~Zhou, Yang~Hu, Weilun~Wang and Houqiang~Li are with the Department of Electrical Engineering and Information Science, 
        University of Science and Technology of China, Hefei, 230027 China 
        (e-mail: cuixiao2001@mail.ustc.edu.cn, zhwg@ustc.edu.cn, eeyhu@ustc.edu.cn, wwlustc@mail.ustc.edu.cn, lihq@ustc.edu.cn).}% <-this % stops a space
\thanks{Corresponding authors: Wengang Zhou and Houqiang Li.}}

% \author{Xiao Cui\\
% CAS Key Laboratory of Technology in GIPAS, EEIS Department,\\
% University of Science and Technology of China\\
% {\tt\small cuixiao2001@mail.ustc.edu.cn}
% \and
%\author{Paper ID: XXXX}

%%%% Figure: Teaser
\twocolumn[{
\renewcommand\twocolumn[1][]{#1}
\maketitle
\begin{center}
  \includegraphics[height=6cm]{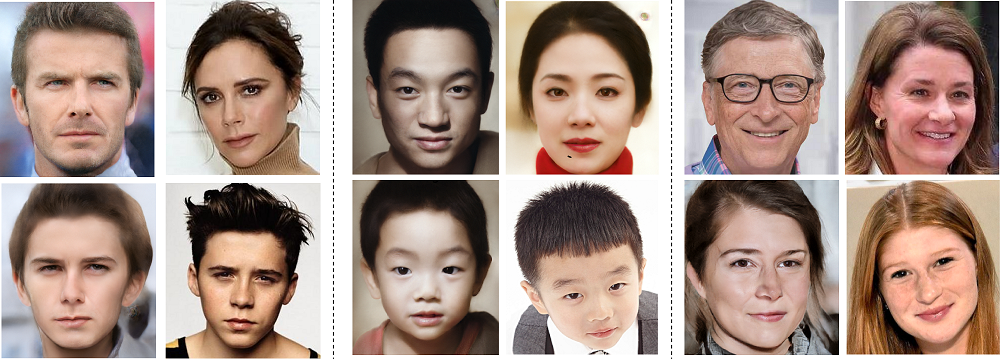}
  %\includegraphics[height=6cm]{pic/front4}
  %\hspace{7pt}
  %\includegraphics[height=6cm]{pic/front6}
  %\hspace{7pt}
  %\includegraphics[height=6cm]{pic/front5}
  \captionsetup{type=figure,font=small}
  \caption{Examples of children images generated by the proposed method. In each group , the images in the first row are the parents, and the second row shows the child generated (left) and the real image of their child (right).}
  \label{front}
  \vspace{2pt}
\end{center}
}]

\blfootnote{Xiao~Cui, Wengang~Zhou, Yang~Hu, Weilun~Wang and Houqiang~Li are with the Department of Electrical Engineering and Information Science, 
        University of Science and Technology of China, Hefei, 230027 China 
        (e-mail: cuixiao2001@mail.ustc.edu.cn, zhwg@ustc.edu.cn, eeyhu@ustc.edu.cn, wwlustc@mail.ustc.edu.cn, lihq@ustc.edu.cn).}% <-this % stops a space
\blfootnote{Corresponding authors: Wengang Zhou and Houqiang Li.}

%%%%%%%%% ABSTRACT
\begin{abstract}

%The Generation Adversarial Network has developed rapidly in image synthesis in recent years and the quality of the generated image has been greatly improved. However, the ability to control and decouple facial features (e.g., eyes, nose, mouth)  is still limited. It is also a new task to use GAN to generate child faces. Using the StyleGAN generator, we introduce a method to generate the corresponding children's pictures according to the pictures of parents under the guidance of biological genetic laws. In this paper, we use large sample gradient method and face landmarks to find the feature vectors corresponding to eyes, nose, mouth, lip, jaw and eyebrow without the help of external annotations, and achieve better disentanglement by weighting irrelevant features and selecting appropriate resolution layer. A novel orthogonalization method is proposed to further decouple the features, so as to precisely control the face attributes of children or parents images with clear semantics. We also adjust the gender features before mixing the parents' images, and achieve the decoupling of the generated images to the background. Our method can go beyond the existing methods and produce satisfactory results.

Generative adversarial networks have been widely used in image synthesis in recent years and the quality of the generated image has been greatly improved. However, the flexibility to control and decouple facial attributes (e.g., eyes, nose, mouth) is still limited. In this paper, we propose a novel approach, called ChildGAN, to generate a child's image according to the images of parents with heredity prior. The main idea is to disentangle the latent space of a pre-trained generation model and precisely control the face attributes of child images with clear semantics. We use distances between face landmarks as pseudo labels to figure out the most influential semantic vectors of the corresponding face attributes by calculating the gradient of latent vectors to pseudo labels. Furthermore, we disentangle the semantic vectors by weighting irrelevant features and orthogonalizing them with Schmidt Orthogonalization. Finally, we fuse the latent vector of the parents by leveraging the disentangled semantic vectors under the guidance of biological genetic laws. Extensive experiments demonstrate that our approach outperforms the existing methods with encouraging results.

\end{abstract}


%%%%%%%%% BODY TEXT
\section{Introduction}
Child image generation aims at synthesizing child face image given the images of  parents. This is a very challenging task since the generated child face should not only resemble the parents, but inherit the attributes following the known genetic laws. Besides, the children born to the same parents may look quite different, which means there is no unique solution to the problem of child image generation. There are many interesting applications on child image generation, such as enabling a couple to preview  the appearance of their children, kinship verification, \emph{etc}.

%Child image generation aims at synthesizing the face of the child given the images of the parents. It can help couples to predict the appearance of their children. It also has important potential applications in kinship verification. This is a very challenging problem since the generated child face should not only resemble the parents, but the inheritance should also follow the known genetic laws. Moreover, since children born to the same parents may look quite different, there is no unique solution to the problem.

There are only a few works studying this child generation problem. KinshipGAN \cite{kinshipgan} uses a deep face network to generate a child's face based on one-to-one relationship. DNA-Net \cite{dnagan} propose to use a deep generative Conditional Adversarial Autoencoder for this task. Although some success has been achieved, those methods suffer three non-trivial issues. First, those methods cannot explicitly control the facial attributes in the generated faces, which significantly limits their application scenarios. Second, the generated images are usually blur and of low quality. Third, they ignore the inheritance law from genetic basis. For instance, thin upper lip is controlled by a dominant gene and is very likely to be inherited. Without considering such prior, the generated child images fail to reflect the inherited attributes.  
 %Ertugrul \emph{et al}. \cite{kin} propose a kinship synthesis framework to generates smile videos of probable children from the smile videos of parents.

%There are only a few works that studing this child generation problem. KinshipGAN  \cite{kinshipgan} proposed to use generative adversarial networks for the task. However, these methods can not be supplemented by the guidance of biological genetic basis: thin upper lip, for example, is a dominant trait and is more easily inherited. The images they generated are quite blur and of low quality. Besides, in these methods, we cannot directly control the facial attributes and specify desired values for them in the generated faces, which is beneficial for the task considering its application scenarios. For example, a user may want to get an image of the child whose eyes are more similar to the mother than to the father.

In the past two years, significant progress have been made on semantic face editing. As an effective tool for editing, latent space is a representation of compressed data where each vector in it corresponds to one orientations. Researchers find that some orientations in the latent space of GANs, such as PGGAN \cite{pggan} and StyleGAN \cite{stylegan}, encode meaningful semantics of human faces. Moving the latent code vectors along these orientation will change the corresponding facial attributes. These results are helpful to develop a child face generator so as to apply genetics knowledge when blending the parents' faces and flexibly control over the attributes of the generated faces. Nevertheless, this idea is difficult to implement as we have to manipulate attributes that are more fine-grained than those edited by previous works.

In this paper, we propose a new framework, \emph{i.e.,} ChildGAN, to generate the face image of the child according to the parents' images under the guidance of genetic laws. The framework consists of two fusion steps: macro fusion and micro fusion. Before these main steps, we first project the parents' images into the pre-trained latent space of StyleGAN. Then after some preprocessing for gender and age alignment as well as backgroud decoupling, we mix the parents' latent codes through a macro fusion module. In order to allow the child to inherit attributes of the parents in a micro way under the guidance of genetic laws, we propose a novel method to identify disentangled sementic directions in the latent space. Our semantic learning method is based on gradient estimation from a large number of samples as well as irrelevant factor reweighting. It finds the important and decoupled semantic vectors in the latent space without the need for manual labels. We also prove that it is feasible to orthogonalize the semantic vectors in the $\mathbb{W}$ space of StyleGAN, which allows conditional manipulation of real images. It is notable that our method does not rely on any external datasets, but only use the images generated by a pre-trained StyleGAN generator and some real images of celebrities to conduct our experiment.

Our main contributions are summarized as follows,

% \begin{itemize}
%     \item We propose ChildGAN to generate children's images based on parents' images. We can control how to inherit from the macro and micro scales.
%     \item By introducing biological genetic law into our generation process, we can produce more realistic images of children.
%     \item We propose a new way to find feature vectors in StyleGAN latent space, which can extract more important and decoupled feature vectors(including the size of eyes, nose, mouth, jaw, eyebrow and the thinkness of lips) without external labels.
%     \item We prove the feasibility of a better orthogonalization method for feature vectors, which can achieve better decoupling between different features, and is more suitable for real image operation.
% \end{itemize}

\begin{itemize}
\item We propose a framework ChildGAN to generate the child's face according to the parents’ faces. Our framework consists of two fusion steps, \emph{i.e.,} macro fusion and micro fusion, which not only integrates the parents’ faces from the holistic perspective, but also involves processing at the micro level.

\item To leverage genetic laws for the generation, a novel method is proposed to disentangle the latent space, which can extract meaningful and decoupled semantic vectors (\emph{e.g.,} semantics corresponding to the size of eyes, nose, mouth, jaw, eyebrows and the thickness of lips) without external labels. 

\item We conduct extensive experiments to verify the effectiveness of our semantic learning module and child generation method. 
    
\end{itemize}

%\item We propose a framework ChildGAN to leverage biological genetic law for generating the face of the child given the parents' images.
%\item A novel method is proposed to disentangle the latent space, which can extract more important and decoupled semantic vectors (\eg, semantics corresponding to the size of eyes, nose, mouth, jaw, eyebrow and the thinkness of lips) without external labels. 
%\item We prove the feasibility of a better orthogonalization method for semantic vectors, which can achieve better decoupling between different features, and is more suitable for real image operation.

\section{Related Work}
In this section Generative Adversial Networks, semantic face editing and face generation are reviewed in detail, since these are three critical ingredients in our face generation approach.

%This paper mainly discusses a special face generation approach based on GANs, and so we summarize the related works for these three aspects.

\subsection{Generative Adversial Networks}
Generative Adversarial Networks (GANs) use the strategy of adversarial training to learn the probabilistic distribution function of training samples for novel sample generation. To date, GANs have been successfully applied to various computer vision tasks, such as semantic attribute editing \cite{semantic1,semantic2,semantic3}, image to image translation \cite{image1,image2,image3,image4} and video generation \cite{video1,video2}. In recent years, many different network structures for GANs have emerged, which significantly improve the stability of the network and the quality of the generated images. For example, PGGAN \cite{pggan} increases the scales of the generator and the discriminator step by step, leading to more stable training and higher quality of generated images. StyleGAN \cite{stylegan} injects style vectors at each convolutional layer with adaptive instance normalization (AdaIN) to achieve subtle and precise style control, which improves the interpretability and controllability of the generation process.
Our work further explores and decouples the latent space of StyleGAN and applies it to child image generation in combination with biological evidence. 

%Generative Adversarial Nets (GANs) use the method of adversarial learning to learn new samples from the training sample. Since GAN was proposed, it has been successfully applied to various computer vision applications, such as semantic attribute editing \cite{semantic1,semantic2,semantic3}, image to image translation \cite{image1,image2,image3,image4} and video generation \cite{video1,video2}. In recent years, DCGAN \cite{dcgan}, InfoGAN \cite{infogan}, PGGAN \cite{pggan}, StyleGAN \cite{stylegan} and other generative adversarial networks have been emerging. The stability of the network and the quality of the generated images are constantly improved, which makes them have great application potential in our life. For example, PGGAN increase the scale of generator and discriminator step by step, which can make training more stable and generate high quality graphs and StyleGAN injects style vectors at each convolutional layer with adaptive instance normalization (AdaIN) to achieve subtle and precise style control, which improves the interpretability and controllability of face generation.

\begin{figure*}
\begin{center}
\includegraphics[width=1\textwidth]{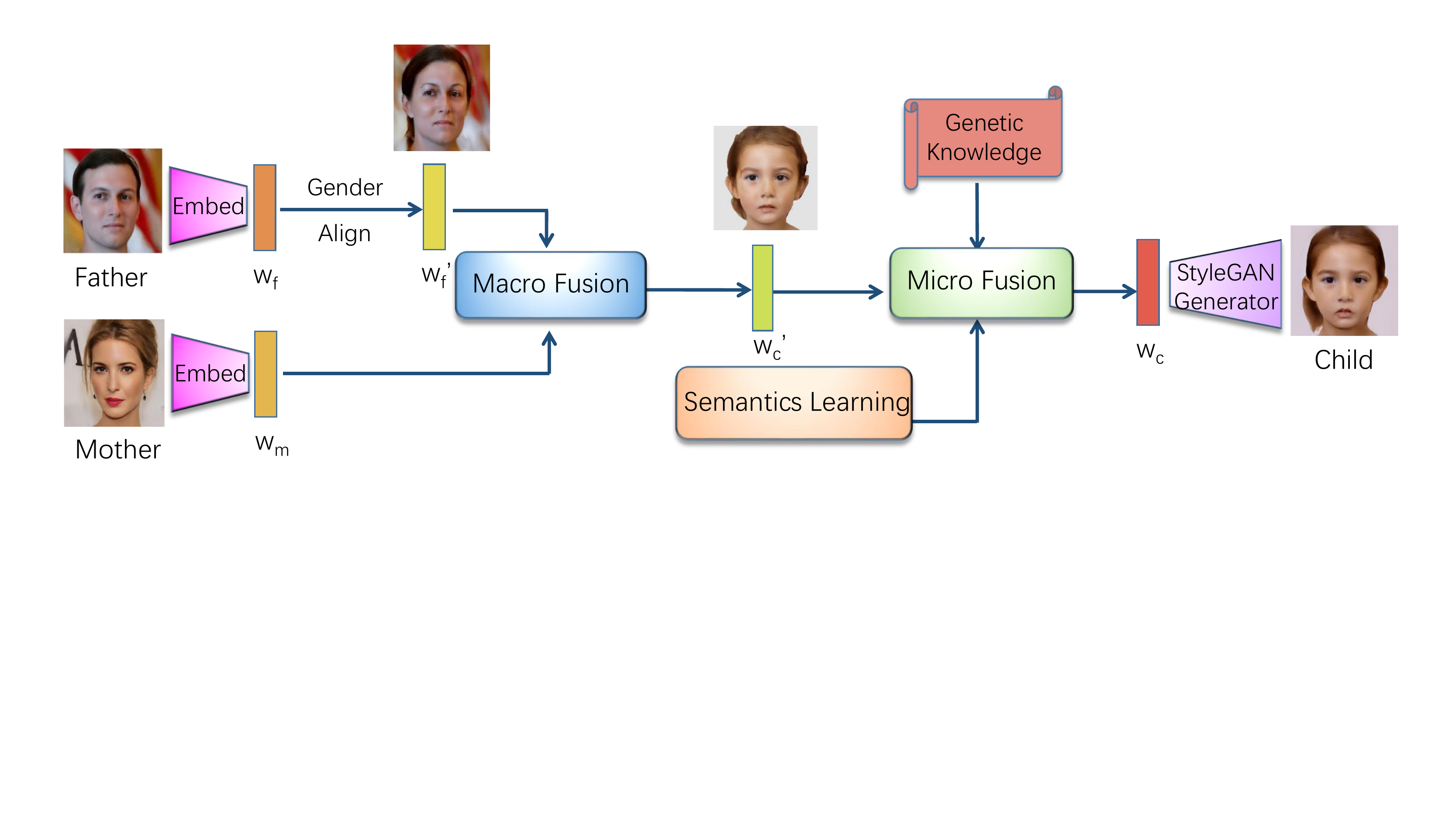}
\end{center}
\vspace*{-4.6cm}
  \caption{The flowchart of our generation process. The images of the parents are first embeded to the latent space of StyleGAN. Then after some preprocessing, the latent codes of the parents are mixed through Macro Fusion. We propose an effective method to identify disentangled semantic directions in the latent space, which allow the child to inherit attributes of the parents in a micro way under the guidance of genetic laws. Finally, the child image is generated from the child's latent code by a pre-tained StyleGAN generator.} 
\label{framework1}
\end{figure*}

%\begin{figure}
%\flushleft
%\begin{center}
%\includegraphics[width=0.45\textwidth]{pic/framework2.png}
%\includegraphics[width=0.45\textwidth]{pic/framework32.png}
%\includegraphics[width=0.45\textwidth]{pic/framework42.png}
%*\includegraphics[width=0.5\textwidth]{pic/2.pdf}
%\end{center}
%\vspace*{-0.8cm}
%\setlength{\belowcaptionskip}{-0.2cm}
%  \caption{The process of micro fusion. The latent codes of parents are projected onto the semantic directions, and the inheritance is carried out under the guidance of genetic laws. }
%\label{framework2}
%\end{figure}

\subsection{Semantic Face Editing}
In semantic face editing, a general idea is to first project the image into some latent space and then edit the latent code. There are two major ways to embed examples from image space into latent space: learning an encoder that maps the given image to the latent space \cite{embed21,embed3}, or starting with a random initial latent code and optimizing it by a gradient descent method \cite{embedder,embed2,embed4,embed5}.

%Researchers embed the image into the latent space and modify the latent code to edit the images. There are two ways to embed examples from image space into latent space: learning the encoder that maps the given image to the latent space \cite{embed21,embed3}, or selecting a random initial latent code and optimizing it by gradient descent method \cite{embedder,embed2,embed4,embed5}.

As for latent code modification, Upchurch Paul \cite{dfi} proved that complex attribute transformation can be realized by linear interpolation in depth feature space. Due to the nice properties of the latent space of StyleGAN, some recent works \cite{interpret,embedder,edit,rig,flow} perform semantic editing based on the StyleGAN model. As a representative method, InterFaceGAN \cite{interpret} trains SVMs to find the directions in the latent space corresponding to attributes including age, gender, pose, expression and the presence of eyeglasses. However, for children image synthesis, we need to edit attributes that are more challenging and fine-grained, such as the size of eyes, nose and mouth. 

%And it is challenging to find the semantics for these attributes with a binary classifier.

%As for latent code modification, Radford et al. found that DCGAN \cite{dcgan} can add the difference between the first two graphs to the third one by increasing the third latent code by the difference between the first two latent vectors. In recent years, some work \cite{interpret,embedder,edit,rig,flow} began to pay attention to StyleGAN's latent space editing. Among them, InterfaceGAN \cite{interpret} found the feature vectors corresponding to age, gender, pose and glasses features. However, their decoupling ability to find the corresponding feature vector is still limited, and they can not find the more critical feature vector which can describe the size of glasses, nose, mouth and so on.

\subsection{Face Generation}
In recent years, great progress has been made with GANs in the field of face generation \cite{mask,attribute,prgan,texture,stylegan,pggan,biggan,stylegan2}. Different from these general face image generation work, we focus on a new interesting problem: kin face generation.

Generating face images of children from images of parents is a relatively new research problem. To our best knowledge, there are only two related works on this topic \emph{i.e.,} KinshipGAN \cite{kinshipgan} and DNA-Net \cite{dnagan}. KinshipGAN \cite{kinshipgan} generates child's face image based on one-to-one relationships (father-to-daughter, mother-to-son, \emph{etc}). The generator follows an encoder-decoder structure, where the encoder extracts the main features of the father or mother, where the decoder uses these extracted features to generate the child's faces. Differently, DNA-Net \cite{dnagan} uses a two-to-one relationship where both images of the father and the mother are used for the generation. However, the model of DNA-Net works mostly in a ``black-box'' fashion. It is difficult to impose knowledge of genetics to guide the generation and control or adjust the results on demand. Besides, the resolution and quality of the images generated by the above two works are worse than those generated by native GANs. In our work, we blend the characteristics of the parents in the latent space of StyleGAN. To make the generation more scientifically sound and controllable, we propose a method to find disentangled semantics in this space, and employ rules of genetics to guide the generation.

%Using GAN to generate images of children from images of parents is still a new field. There are only works on this topic: KinshipGAN \cite{kinshipgan} generates children's faces based on one-to-one relationships (father daughter, father son, mother daughter, and mother son). The generator follows an automatic encoding mode, in which the encoder extracts the main features of the father or mother, and the decoder uses these extracted features to generate children's faces. DNA-Net \cite{dnagan} uses a two-to-one relationship, and the encoder extracts the features of father and mother at the same time. However, synthesis resolution and quality of their models are far behind those of native GANs, and to a large extent, they are in a "black box" state. We can not add the known biological genetic basis to the genetic process, nor can we control the whole genetic process. Unlike them, in our work, we can decouple the features by modifying the StyleGAN latent space, and introduce the known genetic basis into the whole process to make the generation more scientific and interpretable.

\begin{figure}
\begin{center}
\includegraphics[width=0.4\textwidth]{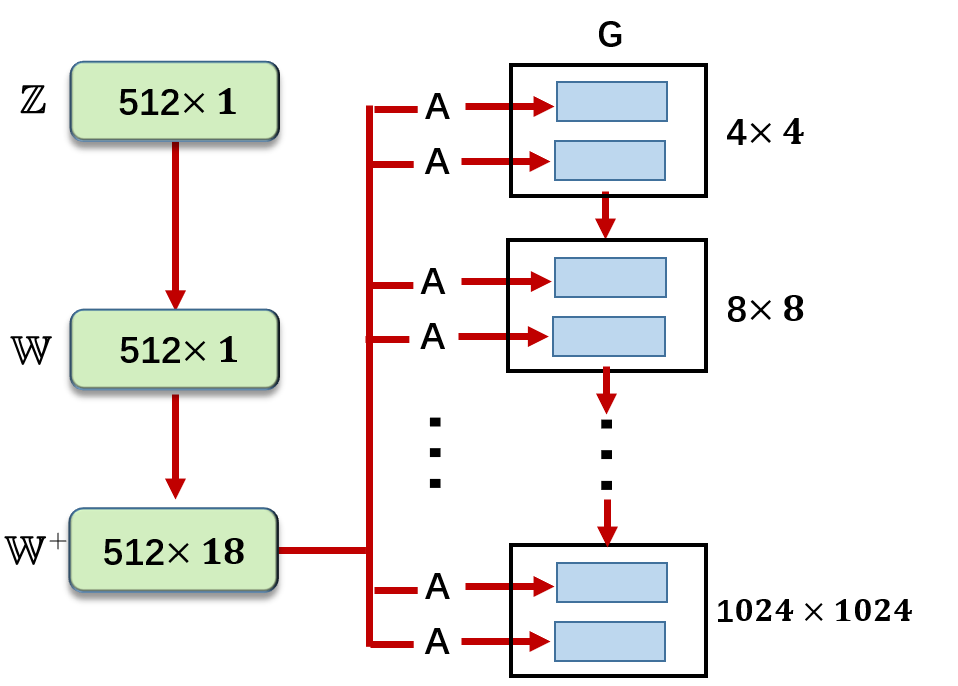}
\end{center}
%\vspace*{-3cm}
  \caption{Framework overview of StyleGAN\cite{stylegan} generator. $\mathbb{Z}$, $\mathbb{W}$ and $\mathbb{W^{+}}$ are three latent spaces, $G$ is the synthesis network, $A$ represents the affine transformation.}

\label{style}
\end{figure} 

\section{Method}
The framework of the proposed ChildGAN is shown in Figure~\ref{framework1} . The face images of the parents are first embeded to the latent space of StyleGAN. Then after some preprocessing, the latent codes of the parents are mixed through Macro Fusion. We propose an effective method to identify disentangled semantic directions in the latent space, which allow the child to inherit attributes of the parents in a micro way under the guidance of genetic laws. Finally, the child image is generated from the child's latent code by a pre-tained StyleGAN generator.  

In this section, we first conduct an overview StyleGAN and Image2StyleGAN, then we introduce the method of macro fusion to achieve the inheritance of macro and relatively rough characteristics. Next we discuss the scientific knowledge for child generation. After that, we introduce the extraction and decoupling of important semantics. Finally, we present how to use the genetic knowledge and orthogonal semantic vectors for micro fusion. 

% In our work, the parents' images are first embedded to the StyleGAN latent space. Then, after age align, the parent's latent codes are mixed through Macro Fusion. We propose an effective method to identify the dissociation semantic direction in the latent space, so that children can inherit their parents' attributes from the micro level under the guidance of genetic law.

\subsection{An overview of StyleGAN and Image2StyleGAN}
As a preparation for the following discussion, we make a brief introduction to StyleGAN \cite{stylegan} and Image2StyleGAN \cite{embedder}. As shown in Figure \ref{style}, in StyleGAN, a non-linear mapping network $f$ first maps a vector $\mathbf{z}\in \mathbb{R}^{512}$ to a vector $\mathbf{w}\in \mathbb{R}^{512}$ in the intermediate space $\mathbb{W}$. After that, we make 18 copies of $\mathbf{w}$ vector and pass them through an affine transformation $A$, whose output is further fed into synthesis network $G$ to control the style of its each layer. There are 18 convolution layers in $G$, two for each resolution, \emph{i.e.,} 9 resolution layers in total. The larger the layer number, the higher the resolution. To use the pre-trained StyleGAN generator, the 18 copies of the $\mathbf{w}$ vector can be extended to 18 different $\mathbf{w}$ vectors whose concatenation constitutes the $\mathbb{W^{+}}$ space. The $\mathbb{Z}$ Space, $\mathbb{W}$ Space and $\mathbb{W^{+}}$ Space are called latent space (latent space specifically refers to $\mathbb{W^{+}}$ space in the following text). We choose to operate on the $\mathbb{W^{+}}$ space because it is suitable for processing real images and has better ability of disentanglement. In order to enable semantic image editing operations to be applied to existing photos, Image2StyleGAN \cite{embedder} proposed an efficient algorithm, where a random initial vector is chosen and optimized using gradient descent, to embed a given image into the latent space of StyleGAN \cite{stylegan}.

\subsection{Macro Fusion}
%In the generator $G$ of StyleGAN, there 18 layers in 9 increasing resolutions.
%w is then copied 18 times, passed through an affine transformation, and then input into a generator G to control the style of its layers

%In a pre-trained StyleGAN \cite{stylegan}, a non-linear mapping network $f$ maps latent space samples $\mathbf{z}\in \mathbb{R}^{512}$ to a vector $\mathbf{w}\in \mathbb{R}^{512}$ in the intermediate latent space $W$ and without considering the noise, the $\mathbf{w}$ vector can completely determine the final image. We choose to operate on $W$ space because it is suitable for processing real images and has better ability of disentanglement.

%By aligning and embedding, we can find the latent code $\mathbf{w}_1$ corresponds to the real father's portrait while  $w_2$ corresponds to the mother's portrait. In order to produce a child with a definite gender (and reduce background clutter), we first need to change the gender characteristics of one of the parents and move the corresponding latent code along the age vector or its opposite direction (we know that moving the dlatent along the feature vector corresponding to a feature can change that feature) to get a new $\mathbf{w}_1$ and $\mathbf{w}_2$. Then we can use the formula $\mathbf{w}=\left( 1-\lambda \right) \mathbf{w}_1+\lambda \mathbf{w}_2$ to make a macro mix.

In order to generate a child's image, we start by making a rough mix of the parents' faces to get a preliminary image of the child. We call this process \emph{macro fusion}. Before our fusion process, we map the parents' images to vectors in the latent space, which called latent codes. Given an image of the father, we first crop and align the face in it. Then similar to Image2StyleGAN \cite{embedder}, we find the optimal latent code $\mathbf{w}_f$ by minimizing the reconstruction loss between the image generated from $\mathbf{w}_f$ and the real image. The latent code $\mathbf{w}_m$ for the mother is obtained in the same way. To produce a child with a specific gender and reduce the background artifacts, we change the gender character of one parent by moving the corresponding latent code along a pre-learned orientation in the latent space which mainly controls the gender attribute. That is to say, if we want the genarated child to be a girl, we need to move the father's latent code vector forward  in this orientation (about 2 units in length) , and if we want the child to be a boy, we need to move the mother's latent code vector backward in this orientation.

%For the father image, first of all, we crop and align the given image to make the face in a relatively fixed position. Then starting from an appropriate initial value, we find the optimal $\mathbf{w}_f$ to minimize the loss function of the similarity between aligned image and the image generated from $\mathbf{w}_f$. Similarly, we can get the latent code $\mathbf{w}_m$ for the mother image. In order to produce a child with a specified gender and reduce background clutter, we need to change the gender characteristics of one of the parents by moving the corresponding latent code along the gender vector \footnote{We use the gender vector extracted by $https://github.com/a312863063/seeprettyface\text{-}face\_editor$}.

%Then starting from a suitable initialization $\mathbf{w}_0$, we search for an optimized vector $\mathbf{w}_f$ that minimizes the loss function that measures the similarity between the aligned image and the image generated from $\mathbf{w}_f$. 

There are two alternatives in macro fusion. First, as a simple method, we can use linear combination: $\mathbf{w}_c'=\left( 1-\lambda \right) \mathbf{w}_f+\lambda \mathbf{w}_m$, where $\lambda$ is a parameter between $0$ and $1$. In this way, every resolution layer feature of the child will be a mixture of the parents. Alternatively, we can control different resolution layers with different $\mathbf{w}$. For example, if we want rough features such as posture, hairstyle, and facial contour to be inherited from the father, while more subtle features such as facial components are inherited from the mother on a macro level, we can set the first two resolution layers be controlled by $\mathbf{w}_f$, while the other layers are controlled by $\mathbf{w}_m$. At the end of the macro fusion, we just need to adjust $\mathbf{w}_c'$ along the pre-learned age vector to generate the child of an expected age.

\subsection{Inheritance Prior for Child Generation}
We adopt some genetic evidence in biology \cite{inherit1} to make our results more scientific (here we only consider Mendelian inheritance). 

%For the following attributes with clear genetic laws, we can use the method to be introduced in Section III.E to let the generated results conform to the biological laws. For other attributes that have not yet been explored by biological researchers, we are not yet able to conduct more scientific operations at the micro level. 

\begin{enumerate}
    \item Skin color \cite{skincolour}: The skin color of a child always follows the natural law of ``neutralizing'' the skin colors of the parents. %For example, if the parents have dark skin, there will never be children with white skin; if one parent is white and the other is black, most of them will give their children a neutral skin color.
    \item Eyes: Big eyes are inherited in a dominant way, so as long as one parent has big eyes, the child is more likely to have big eyes.
    \item Nose: Generally speaking, large, high noses and wide nostrils are in dominant inheritance. If one of the parents has a big nose, it is likely to be inherited by the child.
    \item Jaw: As dominant inheritance, if one parent has a prominent big chin, the child will be more likely to grow into a similar chin.
    \item Lip thickness: A thin upper lip is a dominant inheritance, while a thicker lower lip is a dominant inheritance.
    \item Baldness \cite{bald}: Alopecia is caused by an autosomal dominant gene. Bald men may be heterozygous ($Bb$) or homozygous ($BB$), while bald women are homozygous ($BB$)
\end{enumerate}

If we use $A$ to present dominant gene and $a$ for recessive gene, then the genotype with dominant trait is $Aa$ or $AA$ (in our work, we assume that they're equally likely), and the genotype with recessive trait is $aa$. If one parent presents a recessive trait and the other presents a dominant trait, the probability of the child presenting a recessive trait is: $\frac{1}{2}\times \frac{1}{2}=\frac{1}{4}$ , while the probability of presenting a dominant trait is $\frac{3}{4}$. If both parents display dominant traits, the probability of the child presenting a recessive trait is: $\frac{1}{4}\times \frac{1}{2}\times \frac{1}{2}=\frac{1}{16}$ , while the probability of presenting a dominant trait is $\frac{15}{16}$ . If both parents present recessive traits, their child will surely present a recessive trait. 

In order to classify facial attribute values corresponding to biological traits, we compare the value of the attribute with a threshold value. We get the value for each attribute by by first detecting the face landmarks in the image and then computing the distances between corresponding landmarks. Each threshold is set as the average value of the attribute in a large number of samples. For attributes that cannot be classified by size, we use the projection value of the latent vector on the attribute vector's orientation instead of distance difference as the value of the attribute.

\subsection{Semantics Learning}
As we can see from Section III.B, a genetic rule usually describes how a certain facial attribute is passed down from the parents to the child. To generate the child face according to the genetic laws, we need to identify these attributes in the latent space $\mathbb{W+}$ of StyleGAN, in which we fuse the faces of the parents. However, this is not readily available, since each $\mathbf{w}$ vector usually relates to multiple attributes, \emph{i.e.,} we cannot fuse an attribute by simply interpolating a chosen $\mathbf{w}$. Some previous works \cite{interpret,rig,flow} have shown that there are directions in the latent space of StyleGAN that correspond to different attributes of a face. In this section, we propose an effective method to identify semantic directions (or semantics in short) in the $\mathbb{W+}$ space that separately correspond to the attributes covered by the heredity laws. To ensure that moving a latent vector along one semantic direction affects other attributes as little as possible, we further make these semantic directions orthogonal to each other.

\begin{figure}
\begin{center}
\includegraphics[width=0.42\textwidth]{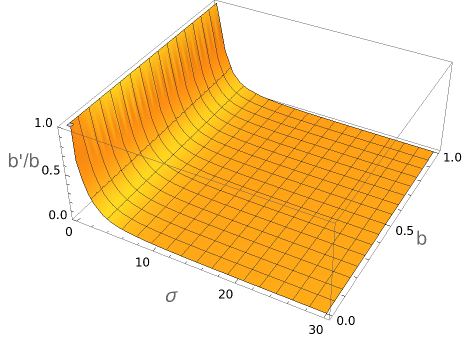}
\end{center}
  \caption{A 3D plot of $b'/b$ at different $b$ and $\sigma$, where $\sigma$ is the variance of $\frac{u_{i,m}-u_{j,m}}{u_{i,l}-u_{j,l}}$, $b$ is the mean of $\frac{u_{i,m}-u_{j,m}}{u_{i,l}-u_{j,l}}$, and $b'$ is the mean of $\mathbf{v}_{l}^{e}$ in the direction of $\mathbf{v}_{m}$ extracted using improved method. In other words, $b$ and $b'$ are the mean values of the components of $\mathbf{v}_{l}^{e}$ in the weighed direction when using the basic method and the improved method respectively.}

\label{mma1}
\end{figure}   

\begin{figure}
\begin{center}
\includegraphics[width=0.45\textwidth]{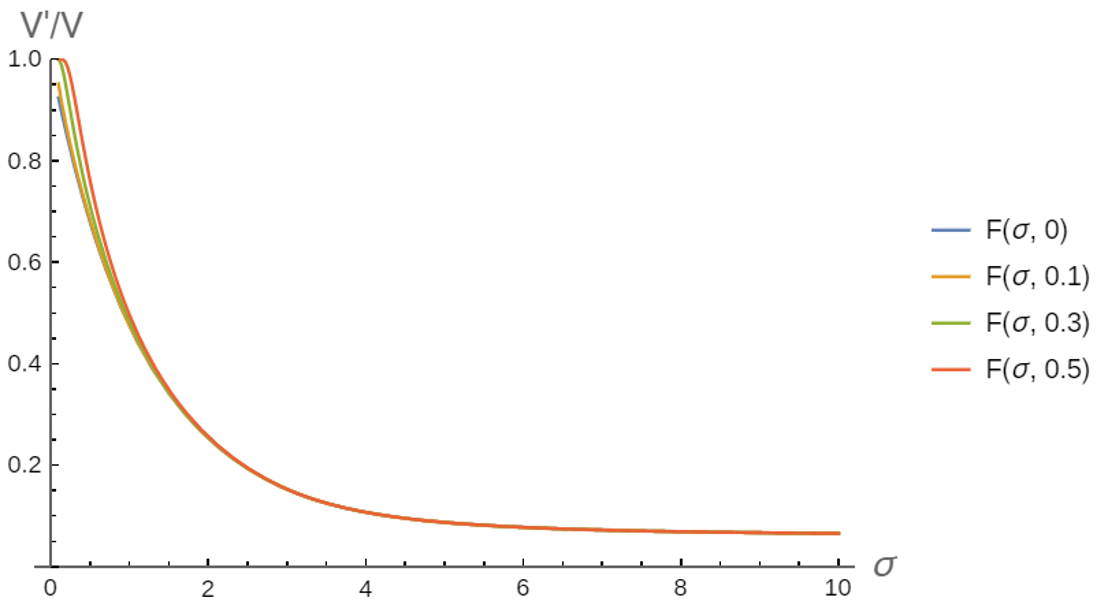}
\end{center}
  \caption{$V'/V-\sigma$ diagram under different $b$, where $b$ is the mean of $\frac{u_{i,m}-u_{j,m}}{u_{i,l}-u_{j,l}}$, $V'$ is the variance of $\mathbf{v}_{l}^{e}$  extracted by our improved method, $V$ is the variance of $\mathbf{v}_{l}^{e}$ extracted by our basic method, and $\sigma$ is the variance of $\frac{u_{i,m}-u_{j,m}}{u_{i,l}-u_{j,l}}$. }

\label{mma2}
\end{figure}   

It is widely observed that we can change the semantics contained in a synthesis continuously by linearly interpolating two latent codes. Let $\{\mathbf{v}_1,\cdots,\mathbf{v}_k\}$ be a set of vectors that we ultimately want, each representing a semantic direction in the latent space. The difference between the latent codes of two images can be represented as a linear combination of the semantic vectors $\{\mathbf{v}_k\}$. The weight for each $\mathbf{v}_k$ is proportional to the change of the value for the corresponding attribute. Without loss of generality, we have:
\begin{equation}
\mathbf{w}_i-\mathbf{w}_j=\sum_k^{}{\left( u_{i,k}-u_{j,k} \right) \times \mathbf{v}_k},
\end{equation}
where $\mathbf{w}_i$ and $\mathbf{w}_j$  are the latent codes of two images. $u_{i,k}$ and $u_{j,k}$ denote the values of the face attribute corresponding to $\mathbf{v}_k$. If $u_{i,k}-u_{j,k}\ne 0$, we have:
\begin{equation}
\frac{\mathbf{w}_i-\mathbf{w}_j}{u_{i,l}-u_{j,l}}=\mathbf{v}_l+\sum_{k\ne l}^{}{\frac{u_{i,k}-u_{j,k}}{u_{i,l}-u_{j,l}}\times \mathbf{v}_k}.
\end{equation}

%If the different features in the image are independent of each other, as the distribution of $\mathbf{u}_l$ is symmetric with respect to 0, we have:

With a large number of image pairs available, we can compute the expectation as:
\begin{equation}
E\left( \frac{\mathbf{w}_i-\mathbf{w}_j}{u_{i,l}-u_{j,l}} \right) =\mathbf{v}_l+\sum_{k\ne l}^{}{E\left( \frac{u_{i,k}-u_{j,k}}{u_{i,l}-u_{j,l}} \right) \times \mathbf{v}_k},
\label{3}
\end{equation}
where $E\left(p \right)$ represents the statistical expectation of $p$. As the distribution of $u_{i,k}-u_{j,k}$ is symmetric with respect to 0, $E\left(u_{i,k}-u_{j,k}\right)=0$, since $u_{i,l}-u_{j,l}\ne 0$, in the same way we have $E\left( \frac{1}{u_{i,l}-u_{j,l}} \right) =0$. So if different semantics are independent of each other, $E\left( \frac{u_{i,k}-u_{j,k}}{u_{i,l}-u_{j,l}} \right) =0$, then we have:
\begin{equation}
E\left( \frac{\mathbf{w}_i-\mathbf{w}_j}{u_{i,l}-u_{j,l}} \right) =\mathbf{v}_l.
\end{equation}

Also we can compute the variance as:
\begin{equation}
\begin{split}
V\left( \frac{u_{i,k}-u_{j,k}}{u_{i,l}-u_{j,l}} \right) = &\quad V\left( u_{i,k}-u_{j,k} \right) V\left( \frac{1}{u_{i,l}-u_{j,l}} \right) \\
&+E^2\left( u_{i,k}-u_{j,k} \right) V\left( \frac{1}{u_{i,l}-u_{j,l}} \right) \\
&+V\left( u_{i,k}-u_{j,k} \right) E^2\left( \frac{1}{u_{i,l}-u_{j,l}} \right), \end{split}
\end{equation}
where $V\left(p \right)$ represents the statistical variance of $p$. According to the above, the last two terms can be dropped because they are 0. So with many sample pairs we have:
\begin{equation}
V\left( \frac{1}{n}\sum_{i,j=i_1,j_1}^{i_n,j_n}{\frac{u_{i,k}-u_{j,k}}{u_{i,l}-u_{j,l}}} \right) =\frac{V\left( u_{i,k}-u_{j,k} \right) V\left( \frac{1}{u_{i,l}-u_{j,l}} \right)}{n}. 
\end{equation}

As we choose to discard the sample pairs with $u_{i,l}-u_{j,l}=0$, the minimum distance difference is 1, $V\left( u_{i,k}-u_{j,k} \right)$ and $V\left( \frac{1}{u_{i,l}-u_{j,l}} \right)$ are finite values, the value of the right side of the above equation goes to zero as $n$ gets very large, that makes:
\begin{equation}
\underset{n\rightarrow \infty}{\lim}V\left( \frac{1}{n}\sum_{i,j=i_1,j_1}^{i_n,j_n}{\frac{u_{i,k}-u_{j,k}}{u_{i,l}-u_{j,l}}} \right) =0.
\end{equation}

So when there are enough samples, we estimate $\mathbf{v}_l$ through:
\begin{equation}
\mathbf{v}_{l}^{e}=\frac{2}{N\times \left( N-1 \right)}\sum_{i=1}^{N-1}{\sum_{j=i+1}^N{\frac{\mathbf{w}_i-\mathbf{w}_j}{u_{i,l}-u_{j,l}}}},
\label{basic}
\end{equation}
where $N$ is the total number of images available for learning, and $\mathbf{v}_{l}^{e}$ is an estimation of $\mathbf{v}_l$.
%Among the formula, $u_{i,l}$ represents the label for the $l$-th feature of the $i$-th picture, $N$ represents the total number of pictures, $\mathbf{d}_i$ represents the $\mathbf{w}$ vector of the i-th picture and $\mathbf{v}_l$ represents the feature vector we desire.

%In order to accelerate the convergence rate and decouple the association between two semantic vectors and reduce the unwanted influence exerted over another attribute when changing the $\mathbf{v}_l$ of an attribute, we add an additional weight to the terms in Equation \ref{basic}:

%In practice, the number of sample pairs is not infinite. In order to speed up convergence and better decouple the association between the two semantic vectors to reduce the unwanted influence exerted over another attribute when changing the $\mathbf{v}_l$ of an attribute, we add an additional weight to the terms in Equation

In fact, the distribution of values of different attributes in sample pictures is not independent of each other. For example, people with larger eyes have larger mouths on average. That makes $E\left( \frac{u_{i,k}-u_{j,k}}{u_{i,l}-u_{j,l}} \right) > 0$ . According to Equation \ref{3}, the vector $\mathbf{v}_{l}^{e}$ we find based on Equation \ref{basic} will have components not only in the $\mathbf{v}_l$ direction, but also in other directions, which can be written as $\mathbf{v}_l+\sum_{k\ne l}^{}{\frac{u_{i,k}-u_{j,k}}{u_{i,l}-u_{j,l}}\times \mathbf{v}_k}$ as the sample size approaches infinity. This means that when we want to change the $l$-th attribute, the rest of the attributes will change as well. In order to reduce the proportion of irrelevant components $\mathbf{v}_{k}\left( k \ne l \right)$ in the extracted vector and reduce the influence of other attributes when changing the $l$-th feature, we add an additional weight to the terms in Equation \ref{basic}:
\begin{equation}
\mathbf{v}_{l}^{e}=\frac{\sum_{i=1}^{N-1}{\sum_{j=i+1}^N{\frac{\mathbf{w}_i-\mathbf{w}_j}{u_{i,l}-u_{j,l}}\times e^{-|\frac{u_{i,m}-u_{j,m}}{u_{i,l}-u_{j,l}}|}}}}{\sum_{i=1}^{N-1}{\sum_{j=i+1}^N{e^{-|\frac{u_{i,m}-u_{j,m}}{u_{i,l}-u_{j,l}}|}}}}
,
\label{improved}
\end{equation}
where $m$ is the index of the semantic being conditioned due to the large value of  $E\left( \frac{u_{i,m}-u_{j,m}}{u_{i,l}-u_{j,l}} \right)$ . By introducing this weight factor, $\mathbf{w}_i-\mathbf{w}_j$ would be weighted heavier if the difference of $m$-th attribute between the two images is small, and $\mathbf{w}_i-\mathbf{w}_j$ would be given a lower weight if the $m$-th attribute of the two images are quite different. This also makes the variance of the restricted directional component smaller and accelerated the convergence speed. Equation \ref{improved} can be further extended if more than one attribute needs to be conditioned. We only need to multiply the weight factors corresponding to these attributes together.

We will prove below that the addition of the weight factor will reduce the expected value of $\mathbf{v}_{l}^{e}$ in the $\mathbf{v}_m$ direction, so that $\mathbf{v}_{l}^{e}$ is closer to the real $\mathbf{v}_l$. Only considering the two-dimensional space composed of $\mathbf{v}_l$ and $\mathbf{v}_m$, Equation \ref{improved} can be rewritten as:
\begin{equation}
\mathbf{v}_{l}^{e'}=\frac{\sum\sum_{i<j}{\frac{u_{i,m}-u_{j,m}}{u_{i,l}-u_{j,l}} e^{-|\frac{u_{i,m}-u_{j,m}}{u_{i,l}-u_{j,l}}|}\mathbf{v}_m+e^{-|\frac{u_{i,m}-u_{j,m}}{u_{i,l}-u_{j,l}}|}\mathbf{v}_l}}{\frac{N\left( N-1 \right)}{2}},
\end{equation}
\begin{equation}
\mathbf{v}_{l}^{e}=\frac{\frac{N\left( N-1 \right)}{2}}{\sum_{i=1}^{N-1}{\sum_{j=i+1}^N{e^{-|\frac{u_{i,m}-u_{j,m}}{u_{i,l}-u_{j,l}}|}}}}\mathbf{v}_{l}^{e'},
\end{equation}
so we can calculate the expectation as:
\begin{equation}
\begin{split}
E\left( \mathbf{v}_{l}^{e'} \right) =&E\left( \frac{u_{i,m}-u_{j,m}}{u_{i,l}-u_{j,l}}\times e^{-|\frac{u_{i,m}-u_{j,m}}{u_{i,l}-u_{j,l}}|} \right) \mathbf{v}_m \\
&+{E\left( e^{-|\frac{u_{i,m}-u_{j,m}}{u_{i,l}-u_{j,l}}|} \right) }\mathbf{v}_l, \\
\end{split}
\end{equation}
\begin{equation}
E\left( \mathbf{v}_{l}^{e} \right) =\frac{E\left( \frac{u_{i,m}-u_{j,m}}{u_{i,l}-u_{j,l}}\times e^{-|\frac{u_{i,m}-u_{j,m}}{u_{i,l}-u_{j,l}}|} \right) }{E\left( e^{-|\frac{u_{i,m}-u_{j,m}}{u_{i,l}-u_{j,l}}|} \right) }\mathbf{v}_m+\mathbf{v}_l.
\end{equation}

Let's assume that $\frac{u_{i,m}-u_{j,m}}{u_{i,l}-u_{j,l}}$ satisfies a normal distribution with mean $0<b<1$ and variance $\sigma ^2 >0$ (it should actually be written as a superposition of a series of different normal distributions, but it makes no difference to the proof). With the weight factor,
\begin{equation}
E\left( \mathbf{v}_{l}^{e} \right) =\frac{\sqrt{\frac{1}{2\pi}}\frac{1}{\sigma}\int_{-\infty}^{\infty}{e^{-\frac{\left( x-b \right) ^2}{2\sigma ^2}-\left| x \right|}}xdx}{\sqrt{\frac{1}{2\pi}}\frac{1}{\sigma}\int_{-\infty}^{\infty}{e^{-\frac{\left( x-b \right) ^2}{2\sigma ^2}-\left| x \right|}}dx}\mathbf{v}_m+\mathbf{v}_l=b'\mathbf{v}_m+\mathbf{v}_l,
\end{equation}
while with the basic method,
\begin{equation}
E\left( \mathbf{v}_{l}^{e} \right) =\frac{\sqrt{\frac{1}{2\pi}}\frac{1}{\sigma}\int_{-\infty}^{\infty}{e^{-\frac{\left( x-b \right) ^2}{2\sigma ^2}}}xdx}{\sqrt{\frac{1}{2\pi}}\frac{1}{\sigma}\int_{-\infty}^{\infty}{e^{-\frac{\left( x-b \right) ^2}{2\sigma ^2}}}dx}\mathbf{v}_m+\mathbf{v}_l=b\mathbf{v}_m+\mathbf{v}_l
\end{equation}

Figure \ref{mma1} shows the relationship between $b'/b$ and $b$ and $\sigma$. We can see that $b'/b$ is less than 1 and approaches 0 when $\sigma$ is large. This indicates that after adding the weight factor, when we try to change the $l$-th attribute, the influence on the $m$-th attribute will be smaller, these two attribute can be better decoupled.

In Figure \ref{mma2}, we calculate the variance ratio of the $\mathbf{v}_{l}^{e}$ extracted by the improved method and the basic method in the $m$-th direction. It can be seen that the addition of the weight factor can significantly reduce the variance, so we only need a relatively small amount of data to make the results converge.

%\begin{equation}
%V\left( \mathbf{v}_{l}^{e'} \right) =\frac{1}{n}\frac{\sqrt{\frac{1}{2\pi}}\frac{1}{\sigma}}\int_{-\infty}^{+\infty}{e^{-\frac{x^2}{2\sigma ^2}-\left| x \right|}}x^2dx
%\end{equation}

%\begin{equation}
%V\left( \mathbf{v}_{l}^{e} \right) =\frac{\frac{1}{n}{\sqrt{\frac{1}{2\pi}}\frac{1}{\sigma}}\int_{-\infty}^{+\infty}{e^{-\frac{x^2}{2\sigma ^2}}}x^2dx}{\left({\sqrt{\frac{1}{2\pi}}\frac{1}{\sigma}}\int_{-\infty}^{+\infty}{e^{-\frac{x^2}{2\sigma ^2}-\left| x \right|}}xdx\right)^2}
%\end{equation}

%\begin{equation}
%V\left( \mathbf{v}_{l}^{e} \right) =\frac{1}{n}{\sqrt{\frac{1}{2\pi}}\frac{1}{\sigma}}\int_{-\infty}^{+\infty}{e^{-\frac{x^2}{2\sigma ^2}}}x^2dx    
%\end{equation}

\begin{figure}
\begin{center}
\includegraphics[width=0.48\textwidth]{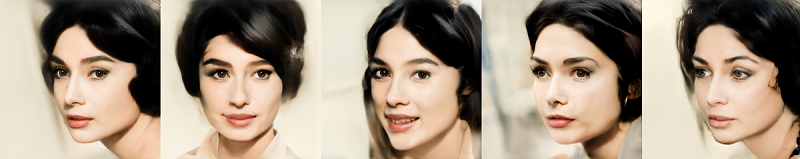}

\includegraphics[width=0.48\textwidth]{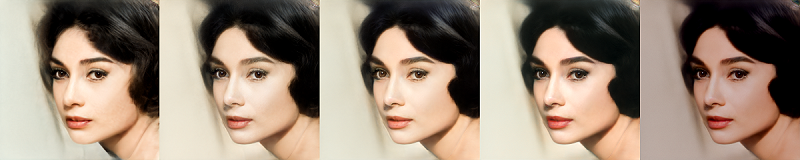}
\includegraphics[width=0.48\textwidth]{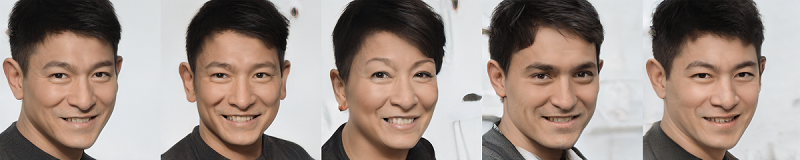}

\includegraphics[width=0.48\textwidth]{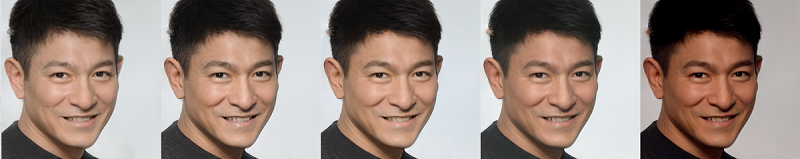}
\includegraphics[width=0.48\textwidth]{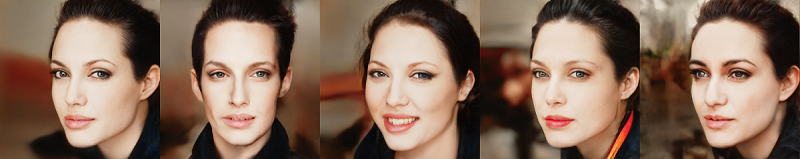}
\includegraphics[width=0.48\textwidth]{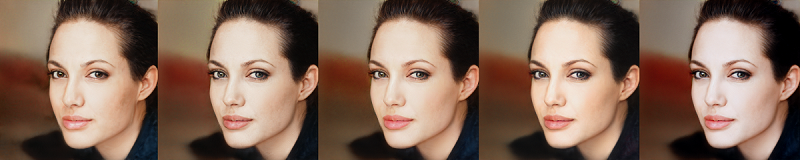}

\end{center}
  \caption{The results of changing the latent codes that control different resolution layers of StyleGAN \cite{stylegan}. For each person, the first picture is the original image, and remaining ones are the results of replacing the latent codes which control the $4\times 4, 8\times 8,\cdots, 1024\times 1024$ resolution layers with a zero vector,respectively. }
\label{fbl}
\end{figure}   

\begin{figure}
\begin{center}
\includegraphics[width=0.49\textwidth]{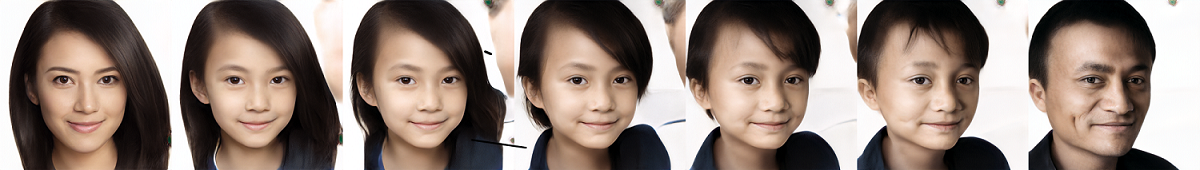}
\includegraphics[width=0.49\textwidth]{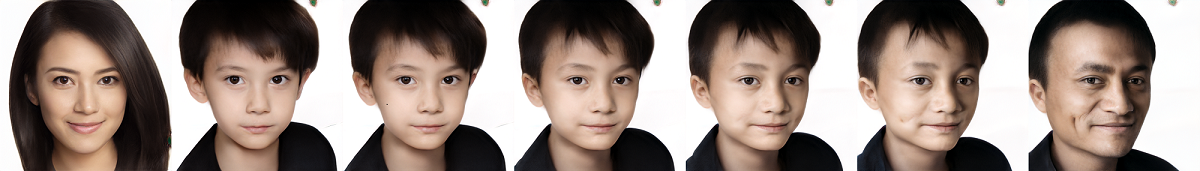}
\end{center}
%\vspace*{-0.5cm}
%\setlength{\belowcaptionskip}{-0.1cm}
  \caption{Examples of gender align. The first and second rows are the results with and without gender align, respectively. }
\label{macro1}
\end{figure}

\begin{figure}
\begin{center}
\includegraphics[width=0.49\textwidth]{macro01.png}
\includegraphics[width=0.49\textwidth]{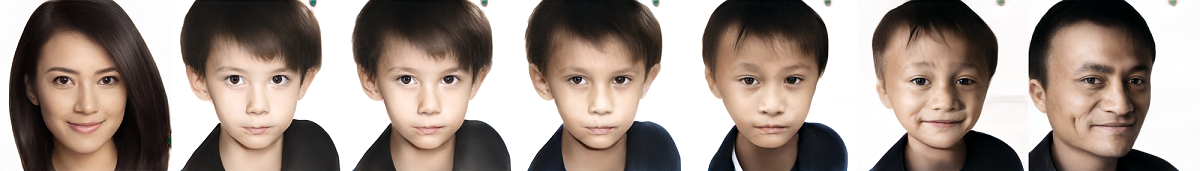}
\includegraphics[width=0.49\textwidth]{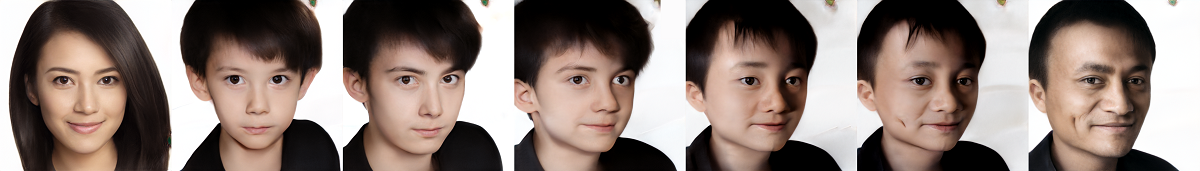}
\end{center}
%\vspace*{-0.5cm}
%\setlength{\belowcaptionskip}{-0.15cm}
  \caption{Examples of macro fusion. The first row is the result of weighted mixing. The second and the third rows are the results of feeding the parents' latent codes to different resolution layers, where in the second row the mother's latent code controls layers with low resolutions and the father's code controls layers with high resolutions and the opposite in the third row. }
  %in the third line the mother's code controls layers with high resolutions and the father's code controls layers with low resolutions.
  %For the ith child image, the resolution layers below the $2\times i$ layer are controlled by the father's code.
\label{macro2}
\end{figure}

\begin{figure*}
\begin{subfigure}[b]{1\textwidth}
\begin{center}
\includegraphics[width=0.48\textwidth]{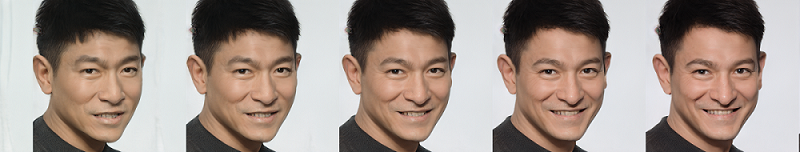}
\includegraphics[width=0.48\textwidth]{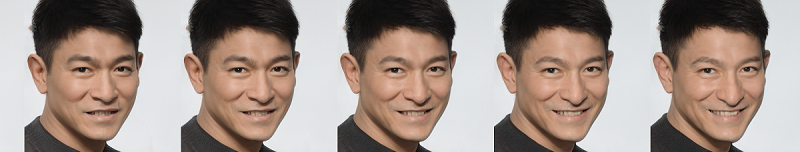}
\vspace*{-0.2cm}
\caption{thickness of upper lip}
\end{center}

\end{subfigure}
\begin{subfigure}[b]{1\textwidth}
\begin{center}
\includegraphics[width=0.48\textwidth]{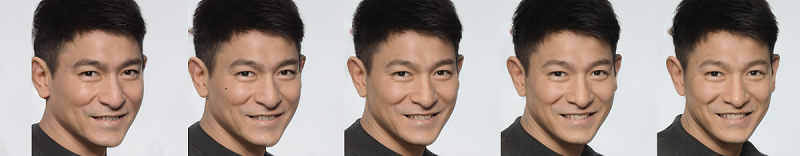}
\includegraphics[width=0.48\textwidth]{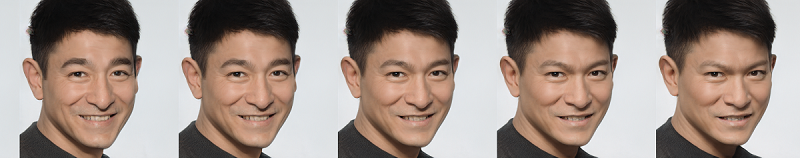}
\vspace*{-0.2cm}
\caption{eyebrow}
\end{center}

\end{subfigure}
\begin{subfigure}[b]{1\textwidth}
\begin{center}
\includegraphics[width=0.48\textwidth]{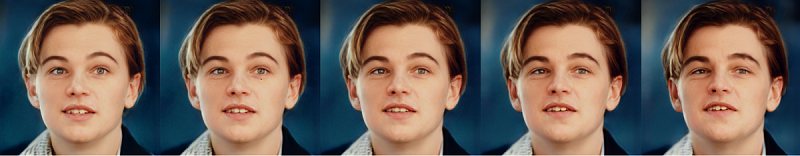}
\includegraphics[width=0.48\textwidth]{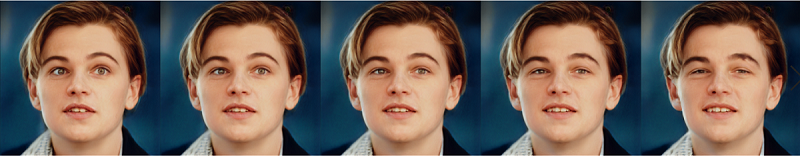}
\vspace*{-0.2cm}
\caption{size of eyes}
\end{center}

\end{subfigure}
\begin{subfigure}[b]{1\textwidth}
\begin{center}
\includegraphics[width=0.48\textwidth]{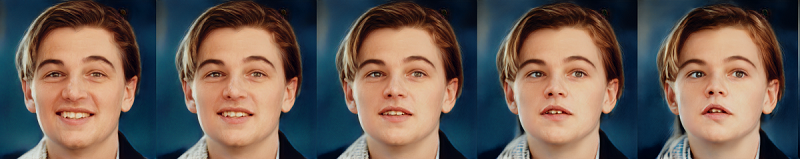}
\includegraphics[width=0.48\textwidth]{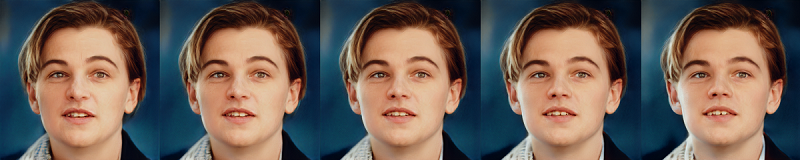}
\vspace*{-0.2cm}
\caption{size of nose}
\end{center}

\end{subfigure}
\begin{subfigure}[b]{1\textwidth}
\begin{center}
\includegraphics[width=0.48\textwidth]{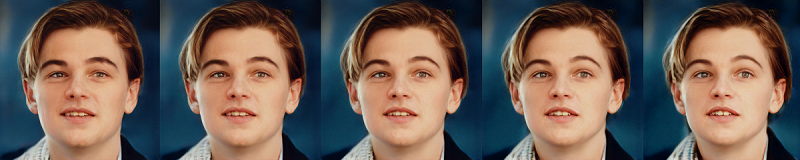}
\includegraphics[width=0.48\textwidth]{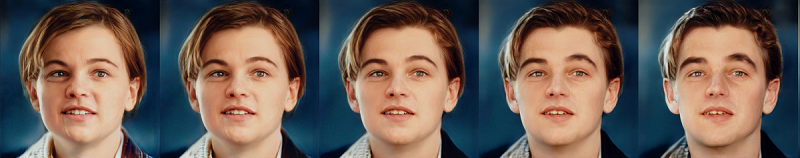}
\vspace*{-0.2cm}
\caption{jaw}
\end{center}
\end{subfigure}
\vspace*{-0.70cm}
  \caption{Single attribute manipulation results. In each row, the five images on the left are the result of using semantic vectors extracted in the basic way, and the five images on the right are the results using semantic vectors extracted in the improved way. For each group of five images, the one in the middle is the original image, and the images on the left and right sides of it are obtained by moving its latent code along the positive and negative directions of the corresponding semantic vector respectively.}
\label{feature}
\end{figure*}
%Through the introduction of this weight factor, when changing the specified feature, the vector which changes the irrelevant feature less will occupy a more important position in the calculation, and the vector with larger change of irrelevant feature will only play a less significiant role, but will not be completely abandoned. If more than one attribute needs to be conditioned, we only need to multiply the corresponding weight factor of these attributes in turn.

In addition to focusing on the overall characteristics of latent space, differences in different resolution layers in latent space facilitate further decoupling. We found that the first resolution layer of StyleGAN mainly controls camera elevation and horizontal angles, while the last four resolution layers mainly control color and background. In order to decouple the extracted facial semantics from attributes that we are not interested in, we can choose to only work on the middle three resolution layers.

The selection of the resolution layers and the estimation of the semantic vectors $\{\mathbf{v}_l\}$  have to a large extent disentangled the semantics. However, there are still some coupling of attributes in $\{\mathbf{v}_l\}$ . To achieve more precise control, we further orthogonalize these vectors. In our work, we use the Gram-Schmidt process to made the semantic vectors orthogonal to each other. Start with $\mathbf{n}_1=\mathbf{v}_1$, we have:
\begin{equation}
\mathbf{n}_l=\mathbf{v}_l-\sum_{i=1}^{l-1}{\frac{\left< \mathbf{v}_l,\mathbf{n}_i \right>}{\left< \mathbf{n}_i,\mathbf{n}_i \right>}\mathbf{n}_i},
\end{equation}
where $\mathbf{v}_l$ is the semantic vector found by Equation \ref{improved} and $\mathbf{n}_i$ represents the orthogonal vector.

In InterFaceGAN \cite{interpret}, the authors failed to decouple some semantics through orthogonalization in the $\mathbb{W}$ space of StyleGAN.  For example, for the ``age'' and ``eyeglasses'' attributes, they found an eyeglasses-included age direction that is somehow orthogonal to the eyeglasses direction itself. So it can not remove eyeglasses from the age direction by orthogonalization. In our experiment, this situation does not happen. In other words, our method can identify semantic directions that are more disentangled than those found by InterFaceGAN. We give an example in Figure \ref{orth1}.

%It has been shown that it is difficult to embed real images into $Z$ space \cite{embedder}. So if we only operate in $Z$ space, we can only modify the images generated by StyleGAN itself. And once we can do orthogonalization in $W$ space, we will be able to modify the real image, which will be very useful.

With the disentangled semantics identified by the method shown in this section, we can decompose the latent vectors of the parents by projecting them onto these semantic directions. Then an attribute of the child will be determined by picking a point in each semantic direction based on the genetic laws. We call this process the micro fusion of the parents.

\subsection{Micro Fusion}
Now that we have obtained the decoupled semantic vectors that correspond to key attributes of the face, we can inherit face components of the parents according to the genetic laws. Based on the preliminary child latent code obtained after macro fusion, we further adjust it in the semantic directions. For each semantic vector, we first project the parents' and preliminary child's latent codes onto it. For the case that one parent presents a dominant trait while the other parent presents a recessive trait, we get the child's phenotype according to probability, and move the child's latent code to the father or mother's projection. If both parents show dominant traits but the child should show the recessive character according to probability, we move the child's latent code across the less obvious dominant side and move on until it becomes recessive. In the case of parents and child all showing dominant traits, we make the child's latent code move randomly under the restriction of parents' projection in this direction (the same for both parents with recessive traits or when there is no clear genetic rule to guide the semantic). 

After dealing with each semantic direction in accordance with the above methods, we resynthesize $\mathbf{w}_c$ by:
\begin{equation}
   \mathbf{w}_c=\mathbf{\hat{w}}_c’+\sum_l{p_l\mathbf{v}_l},
\end{equation}
where $\{\mathbf{v}_l\}$ are the semantic vectors, $p_l$ is the projection component on each semantic direction and $\mathbf{\hat{w}}_c'$ is what's left of $\mathbf{w}_c'$ after been decomposed. After that, we send $\mathbf{w}_c$ into the StyleGAN Generator to obtain the final child image.

%If there is no clear genetic rule to guide the semantic, the child's latent code will be randomly moved in this direction under the restriction of the latent code projection of both parents.

%After finding the feature vectors corresponding to different features and decoupling the latent space, we can inherit each feature according to the genetic law. For each feature, we first project the father's and mother's latent codes onto the direction of the feature vector. If the feature is guided by a clear genetic law, the mixed latent code from Sec. 3.1 will be moved to the father's side or the mother's side according to the probability. If there is no clear genetic rule to guide the feature, the mixed latent code will be randomly moved in this direction under the restriction of the latent code projection of both parents.

\begin{figure}[htbp]
\begin{center}
\includegraphics[width=0.4\textwidth]{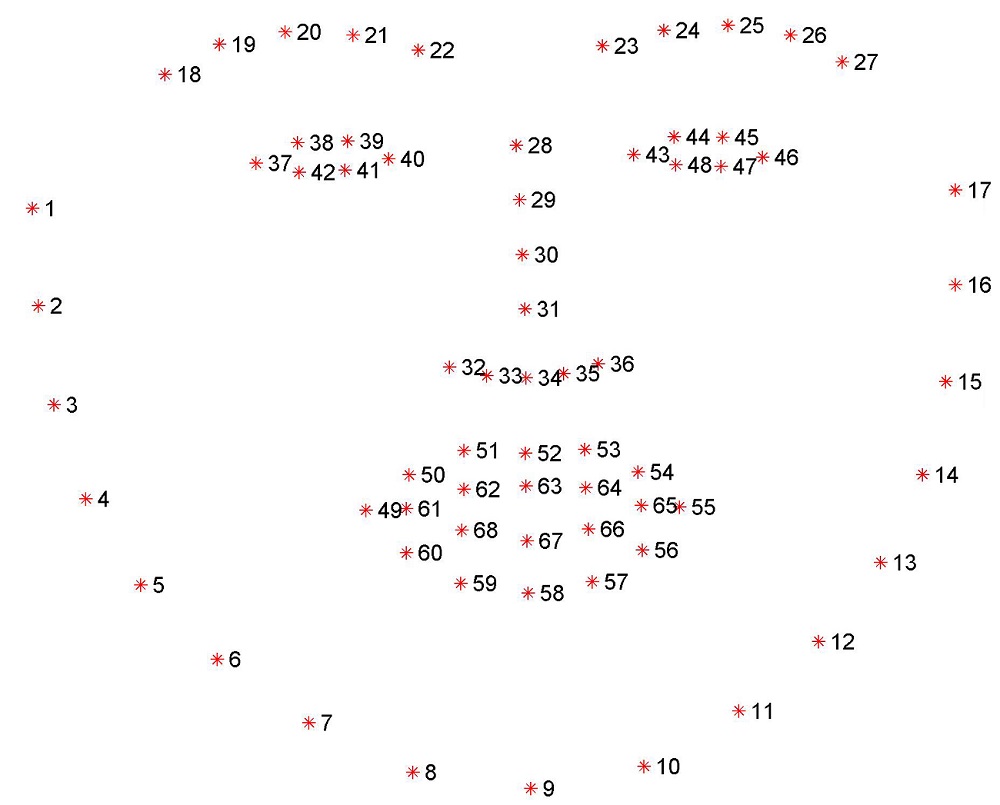}
\end{center}
  \caption{Positions of the 68 landmarks on a human face. \protect\footnotemark[1]}
\label{landmark}
\end{figure}

\begin{table}[htbp]
\begin{center}
\begin{tabular}{|l|c|}
\hline
 Attribute & Index of landmarks \\
\hline\hline
 length of eyebrow & $\left(18,22\right),\left(23,27\right)$ \\
 width of eyes & $\left(38,42\right),\left(45,47\right)$ \\
 length of eyes & $\left(37,40\right),\left(43,46\right)$ \\
 width of nose & $\left(34,36\right),\left(32,34\right)$ \\
 thickness of upper lip & $\left(52,63\right),\left(51,62\right),\left(53,64\right)$ \\
 thickness of lower lip & $\left(58,67\right),\left(59,68\right),\left(57,66\right)$ \\
 width of mouth & $\left(52,58\right),\left(51,59\right),\left(53,57\right)$ \\
 length of mouth & $\left(49,55\right)$ \\
 chin shapeness & $\left(8,10\right),\left(7,11\right)$ \\
\hline
\end{tabular}
\end{center}
\caption{The face attributes we consider and the corresponding landmark pairs we use to get the labels for the attributes. }
\label{landmarktable}
\end{table}

%We use the coordinate difference between different landmarks to represent the size of different facial features. Thus, we describe the size and thickness of most of the main features of the face. We use the coordinate difference between different landmarks to represent the size and thickness of different facial features,so we do not need to use external labels or manual marking.
\section{Experiments}
   
We use the extended $\mathbb{W}$ space in StyleGAN, namely $\mathbb{W}^+$ space, which is a concatenation of 18 different 512-dimensional $\mathbf{w}$ vectors, to carry out our experiment. Previous works \cite{embedder} have shown that the $\mathbb{W}^+$ space is more effective than the original $\mathbb{W}$ space. In this section, we will first reveal the influence of gender alignment and two kinds of macro fusion, and then show the effects of semantics learning for micro fusion, including the capability of different extraction methods for semantic vectors and the effectiveness of orthogonalization in $\mathbb{W}^+$ space. After that, we will present the results of child generation and show how the genetic laws work. Finally we will present some quantitative and visual results.

\begin{figure}
\begin{center}
\includegraphics[width=0.48\textwidth]{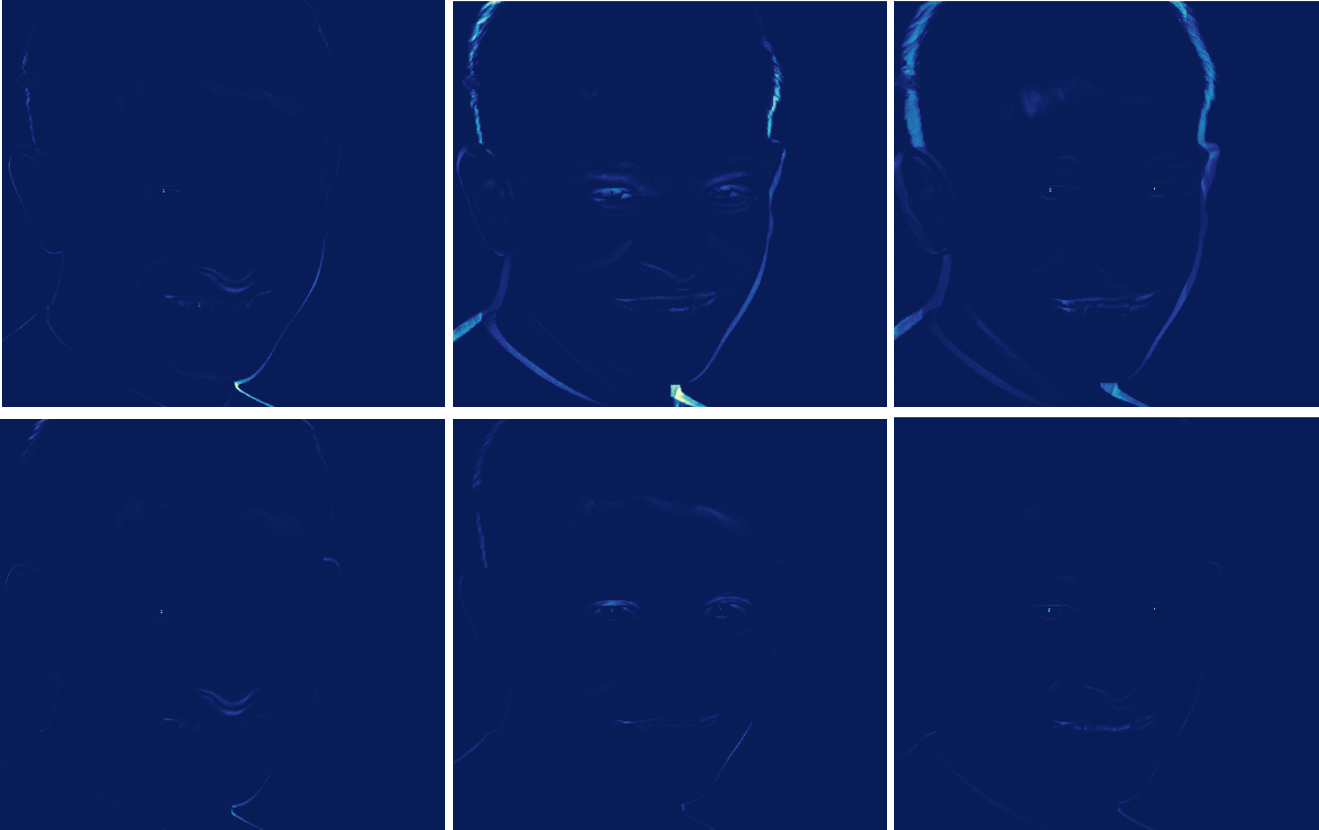}
\end{center}
%\vspace*{-0.4cm}
%\setlength{\belowcaptionskip}{0.2cm}
  \caption{Heat maps of the mean squared error between the edited outputs and original image. The first and second rows are the results of using the basic method and the improved method, respectively. The edited attributes are the nose size (left), eye size (middle) and upper lip thickness (right).}
\label{heatmap}
\end{figure}

\begin{figure}
\begin{center}
\includegraphics[width=0.4\textwidth]{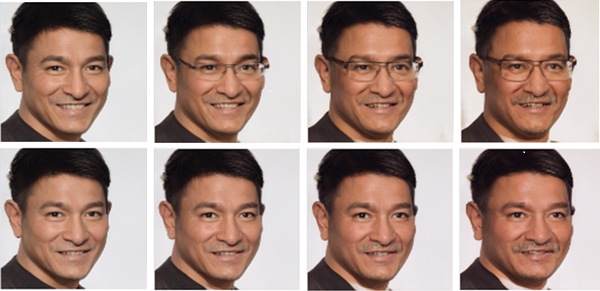}
\end{center}
%\vspace*{-0.5cm}
%\setlength{\belowcaptionskip}{-0.3cm}
  \caption{Analysis on whether it is effective to do orthogonalization in $\mathbb{W}$ Space of StyleGAN \cite{stylegan}. The first row and the second row are the results of aging without and with orthogonalization, respectively. }
  %(and eyeglasses vector is placed before age vector in the process of orthogonalization).
\label{orth1}
\end{figure}

%\begin{figure}
%\begin{center}
%\includegraphics[width=0.4\textwidth]{pic_lr/orth1}
%\includegraphics[width=0.4\textwidth]{pic_lr/orth2}
%\includegraphics[width=0.48\textwidth]{pic/nose21.png}
%\includegraphics[width=0.48\textwidth]{pic/nose22.png}
%\includegraphics[width=0.48\textwidth]{pic/nose4.png}
%\includegraphics[width=0.48\textwidth]{pic_lr/LDnose1.png}
%\includegraphics[width=0.48\textwidth]{pic_lr/LDornose.png}
%\includegraphics[width=0.48\textwidth]{pic_lr/LDnoseorth2.png}
%\end{center}
%\vspace*{-0.55cm}
%\setlength{\belowcaptionskip}{-0.6cm}
%  \caption{An example of the orthogonalization operation. The first row and the second row are the results of using and not using the orthogonalization process after extracting the semantic vector the correspond to the size of nose with our basic method. }
  %As you can see, changing the size of the nose in the first line is accompanied by a change in the smile, while in the second and third line the problem is greatly alleviated.
%\label{orth2}
%\end{figure}

\begin{figure*}
\begin{center}
\includegraphics[width=0.95\textwidth]{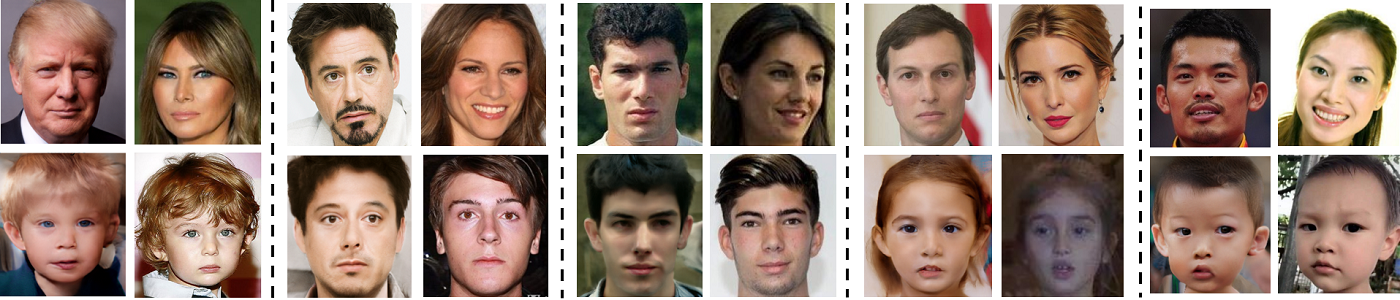}
\end{center}
%\vspace*{-0.55cm}
%\setlength{\belowcaptionskip}{-0.25cm}
\caption{More results of child generation. In each group, the images in the first row are the parents, while the second row shows the child generated by the proposed method (left) and the real image of the child (right)}
\label{result}
\end{figure*}

\begin{figure}
%\flushleft
\begin{center}
%\centering
\includegraphics[width=0.42\textwidth]{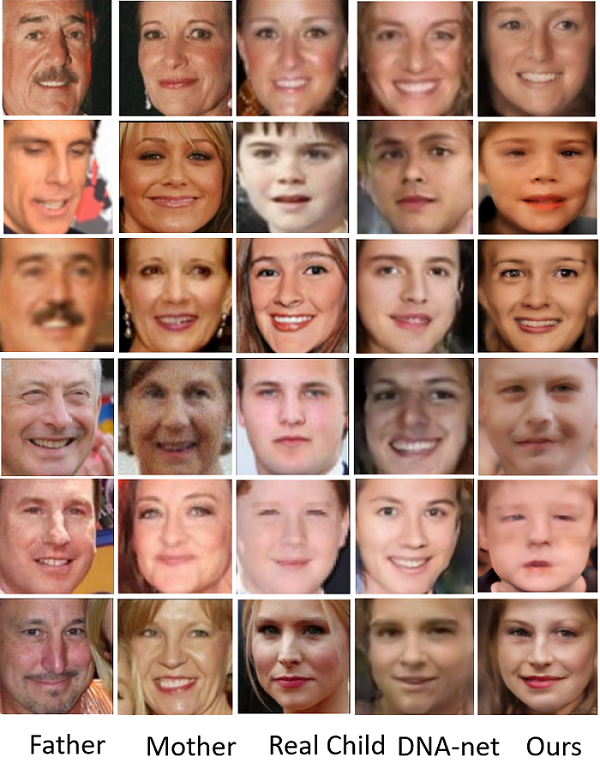}
\end{center}
%\vspace*{-0.7cm}
%\setlength{\belowcaptionskip}{-0.1cm}
  \caption{Comparison between our results and the results obtained by DNA-Net \cite{dnagan}. }%Each line, from left to right, is the real image of the father, the real image of the mother, the real image of the child, our results, the results shown in the DNA-net paper
\label{compare}
\end{figure}

\begin{figure}
\begin{center}
%\centering
%\flushright
\includegraphics[width=0.42\textwidth]{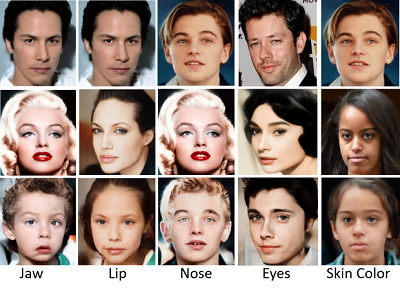}
\end{center}
%\vspace*{-0.8cm}
%\setlength{\belowcaptionskip}{-0.15cm}
\caption{Examples of the role of genetic laws. Each column is the result of considering one genetic factor, and all children inherit the corresponding dominant traits. }
\label{inherit}
\end{figure}

\begin{figure}
\begin{center}
\includegraphics[width=0.32\textwidth]{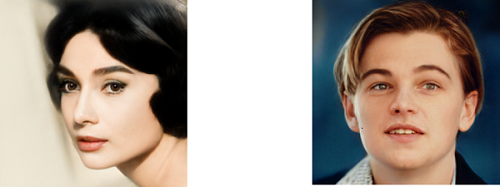}
\includegraphics[width=0.42\textwidth]{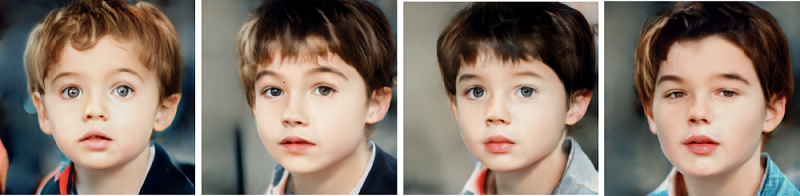}
\includegraphics[width=0.42\textwidth]{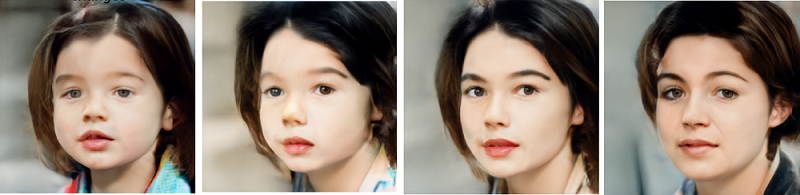}
\end{center}
%\vspace*{-0.5cm}
%\setlength{\belowcaptionskip}{-0.3cm}
\caption{An example about the diversity of generated results. The first row shows the images of the parents, the second and the third rows are the results of children with different genders, ages, and genetic patterns. }
\label{variety}
\end{figure}

\begin{figure}
\begin{center}
\includegraphics[width=0.38\textwidth]{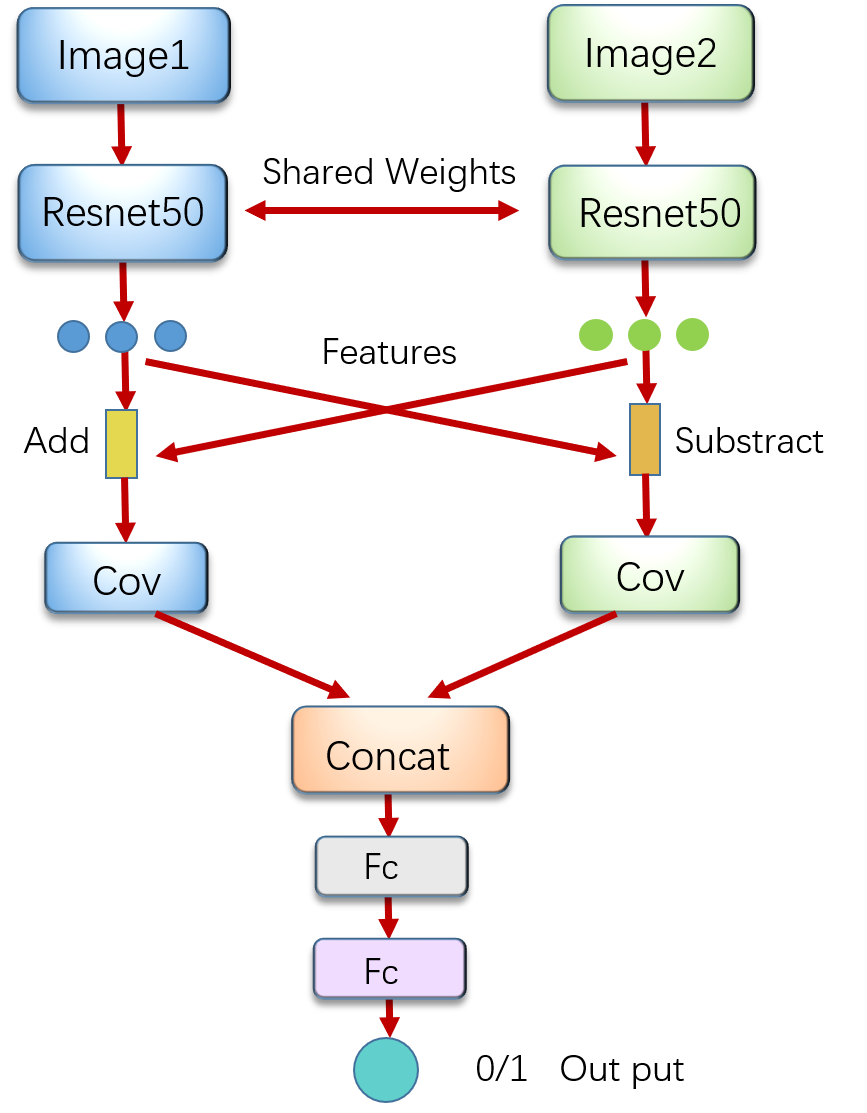}
\end{center}
%\vspace*{-0.5cm}
%\setlength{\belowcaptionskip}{-0.1cm}
  \caption{Kinship vertification network architecture.}
\label{vertification}
\end{figure}

\begin{figure}
\begin{center}
\includegraphics[width=0.4\textwidth]{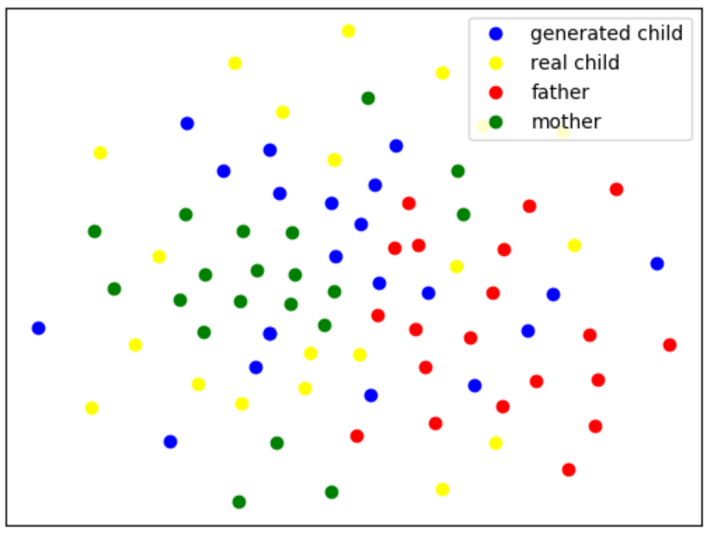}
\end{center}
%\vspace*{-0.5cm}
%\setlength{\belowcaptionskip}{-0.3cm}
\caption{Visualization of facial feature distribution of fathers,mothers, children, and generated ones. Red points represent the feature of fathers, green for mothers, yellow
for real children, blue for generated children, respectively. Best viewed in color.}
\label{tsne}
\end{figure}

\subsection{Evaluation of Macro Fusion}
%As shown in Fig. \ref{macro}, in the process of mixing parents' latent codes with different resolution levels or mixing all the latent codes are in proportion directly, if the above two latent codes are mixed without any processing, the background of the generated image will be disordered, and if the gender characteristics of one of the parents is adjusted before the above-mentioned mixing, the background of the image will not appear disorderly block spots. This shows that there is coupling between gender features and background features in StyleGAN's $W$ space. For different genders, the latent codes corresponding to the same background features are different. In order to remove the background confusion, we need to adjust the gender characteristics of one of the parents before mixing the macro characteristics of parents and adjusting the age to produce children.
We explore the effects of the latent codes for different resolution layers by replacing each latent code with a zero vector, respectively. As shown in Figure~\ref{fbl}, the code for first resolution layer mainly affects the face orientation and elevation angle, the latent codes for the last four resolution layers mainly affect the color and illumination. That enables us to decouple these irrelevant attributes by keeping the resolution layers mentioned above unchanged. 

Figure~\ref{macro1} shows that there is coupling between gender and background in StyleGAN's $\mathbb{W}$ space. Under different gender attributes, the same background corresponds to different latent codes, resulting in background artifacts after macro fusion. In order to remove the background confusion, we adjust the gender of one of the parents before the macro fusion.

Figure~\ref{macro2} shows the results of macro fusion using the two different ways respectively. In the case of a weighed combination, with the increase in the proportion of father's latent code, different attributes of the child are equally shifted from mother-like to father-like. As for feeding the parents' latent codes to different resolution layers, it can be seen that in the second row, from left to right, the child's subtle features first change from maternal to paternal, and it is only after the subtle changes are complete that the rough features begin to shift to paternal. As opposed to this, in the third row the child's rough features first change from mother-like to father-like while the subtle features change later.

\footnotetext[1]{\label{note1} From $ $ $http://dlib.net/files/shape\_predictor\_68\_face\_land$ $marks.dat.bz2$}

%(the first 2$i$ resolution layers being controlled by the father in the $i$th figure)
%(if we want to the child we generate to be a boy, we should change the mother's latent code to make its gender characteristics become male).

%(for example, the first two layers of latent codes are provided by the father, and the later are provided by the mother)

%(move the latent code of one parent along the feature vector corresponding to the gender feature)
\subsection{Evaluation of Micro Fusion}

We use 10,000 images generated randomly by the pre-trained StyleGAN to learn the semantic vectors. We compare two versions of our learning method. In the basic method, we use Equation \ref{basic} for semantics extraction. In the improved version of our method, we use Equation \ref{improved} and keep the first two and the last six $\mathbf{w}$ vectors in the $\mathbb{W}^+$ space unchanged. As we use the coordinate difference between different landmarks to represent the size or thickness of different facial semantics, we do not need to use external labels or manual markings. 

%The positions of the 68 landmarks on human face and the landmarks selected to measure the sizes of the facial components are shown in the supplementary file.
For a given face image, we get the labels for the attributes, such as the size of the eyes, the size of the nose, and the thickness of the lip, by first detecting the face landmarks in the image and then computing the distances between corresponding landmarks pairs. We use a detector \footref{note1} which can locate 68 keypoints of face components. The distances between pairs of landmarks are used as quantitative labels for the attributes.

In Figure~\ref{landmark} we show the positions of the 68 landmarks on human face. The landmark pairs selected to measure the sizes of the facial components are displayed in Table~\ref{landmarktable}. By calculating the distance between each pair of landmarks, we describe the size or thickness of the main components of the face without using external labels or manual marking. For some attributes, more than one set of landmark points can be used to describe them. We take the average of distances to get the corresponding labels. For example, we use the average of the distances between $\left(38,42\right)$ and between $\left(45,47\right)$
as a label for the width of eyes.

%We use 10000 randomly generated images to extract feature vectors according to the basic method, and use 10000 randomly generated images with the first two and the last six dimensions of the $\mathbf{w^+}$ vector unchanged to extract the feature vectors according to our improved method.

%From the heatmaps shown in Fig. \ref{heatmap}, we can see that compared with our basic method, our improved method can focus on the object of interest better, and will no longer present the face contour (this shows that the feature vector extracted by the improved method will not modify the face contour that we do not want to adjust).

Figure~\ref{feature} shows that with the semantic vectors extracted by the basic method, we are able to control the important facial attributes, but there are still some coupling problems with other irrelevant attributes such as the change of the pose. With the improved method, these problems have been greatly alleviated.  As shown from the heatmaps in Figure~\ref{heatmap}, our improved method focuses on the component of interest better than our basic method, and it no longer modifies the face contour. These all demonstrate that we can manipulate the facial attributes better with less attribute coupling.

%\subsubsection{Can orthogonalization of W-Space work?}

To demonstrate the effect of the orthogonalization operation, like InterFaceGAN \cite{interpret}, we first tested the coupling of age and eyeglass attributes in Figure~\ref{orth1}. We observe that the appearance of glasses will be accompanied by the increase of age. However, after the orthogonalization process, the age attribute will no longer be coupled with the eyeglasses semantic. Also, the cosine similarity between the original age vector and the eyeglass vector is 0.13, which is not a very close number to 0. All these show that the orthogonalization of age vector is effective. Other attributes also show this effectiveness. In addition, we find that the order of the semantic vectors in the process of orthogonalization has little effect on the decoupling results, which demonstrates the robustness of the operation.

\begin{table}
\begin{center}
\begin{tabular}{|l|c|c|}
\hline
Item & Accuracy(Epoch=10) & Accuracy(Epoch=20)\\
\hline \hline
FIW test data & 0.696 & 0.729 \\
Gene(ours) & 0.739 & 0.870 \\
Real & 0.714 & 0.722\\
Gene(DNA-Net) & 0.563 & 0.563 \\
\hline
\end{tabular}
\end{center}
\caption{Kinship verification scores. The first row is the accuracy of the network's identification of kinship in the FIW \cite{FIW} DataSet, and the following each row is the proportion of the corresponding child-parent pairs judged by the network to have kinship among all the pairs}
\label{vertification2}
\end{table}

\begin{table}
\begin{center}
\begin{tabular}{|l|c|}
\hline
Type & Average distance \\
\hline\hline
Real - real pairs & 1.194 \\
Gener - real(randomly selected) pairs & 1.189 \\
Gener - real(corresponding) pairs & 0.943\\
\hline
\end{tabular}
\end{center}
\caption{Similarity scores computed between every two features extracted by Facenet using cosine distance}
\label{similarity}
\end{table}

\subsection{Results of Child Generation}
Figure~\ref{front} and Figure~\ref{result} show some of our generated results. As can be seen, all facial attribues of children are combinations of the corresponding parents. The children's attributes may be biased to one of the parents, which is mainly regulated by the genetic rules. Figure~\ref{compare} compares our method with DNA-Net \cite{dnagan}. Although the input images given by their paper have low resolution, our results still show great advantages. In Figure~\ref{inherit}, we show how the known genetic laws act on specific characteristics. The results demonstrate that the dominant traits of parents are more likely to be passed on to their children, which makes our generation process more scientifically sound and reliable.

%Fig. \ref{result} shows some of our generated results, which are quite encouraging. Fig. \ref{compare} shows that we used some images the same as DNA-net to generate results and compared them with the results shown in that paper. Although the input image resolution was extremely low and the StyleGAN's advantages could not be well played, our results still showed great advantages.

In Figure~\ref{variety} we show the diversity of our generated results. We can generated children with different ages and genders. Also, since the heredity of various characteristics follows the Mendelian inheritance law, it is not deterministic but with a certain probability that different children of the same parents do not look exactly the same in appearance. Our results show this diversity. 

\subsection{Quantitive Evaluation}
We quantitatively evaluate the proposed method through kinship vertification and similarity calculation, and visualize the feature distribution.

For kinship vertification, we used two Resnet50 \cite{resnet} pre-trained on the VggFace dataset \cite{vggface} to extract image features respectively and fine-tune the entire network on the FIW \cite{FIW} dataset. The structure of network is shown in figure~\ref{vertification}. For the network, given two pictures of faces, it can output whether the two people in the pictures are related. After 10 and 20 rounds of training, 69.6$\%$ and 72.9$\%$ accuracy were achieved on the FIW \cite{FIW} data set respectively, and similiar performance was also achieved in other real image samples we selected, indicating that no over-fitting occurred.

We submitted the generated child images and the images of their corresponding parents to the network for judgment. The higher the proportion of the groups considered by the classifier to have kinship, the better the performance of the generation method. As shown in Table~\ref{vertification2}, in our generation method, the kinship vertification network trained for 10 and 20 epochs considers 73.9$\%$ and 87.0$\%$ of image pairs to have kinship relations respectively, which is close to or even better than the results of real image pairs. And images generated by DNA-Net \cite{dnagan} achieved  56.3$\%$ and 56.3$\%$, respectively.

We use a pre-trained Facenet \cite{facenet} to perform the similarity test. Facenet \cite{facenet} extracts the image identity features and directly learns the mapping between the image and the points on the Euclidean space of the feature. The distance of the points corresponding to the two images directly corresponds to the similarity of the two images. The training data are totally independent from FIW. As shown in Table~\ref{similarity}, the average distance of real-to-generated pairs is 0.943, compared to 1.189 for generated faces and random real faces and 1.194 for real-real pairs. This indicates that the child images generated by us have good feature similarity with the corresponding images of parents.

T-Distributed Stochastic Neighbor Embedding \cite{tsne} is a machine learning algorithm for dimensional reduction. We use it to reduce the dimensionality of high-dimensional facial features to 2-dimensional for visualization. As shown in Figure~\ref{tsne}, unlike in DNA-Net \cite{dnagan} where the features the generated children are concentrated on one side and far away from real ones, the features of children generated by us are evenly distributed. The distribution of image features generated by us is similar to that of real children's images, and the generated image features are also very close to parents' image features (some are closer to father's features and some are closer to mother's features), which is consistent with our vertification results.

\section{Conclusion}
We proposed ChildGAN to generate child images from the images of a couple under the guidance of genetic laws. Based on two fusion steps,our approach not only integrates the parents’ faces from the macro perspective, but also processes at the micro level. With a new method for semantic learning, we can can precisely and independently control key facial attributes including eyes, nose, jaw, mouth and eyebrows. Extensive experiments suggest that our semantic learning module and child generation method are effective. For further work, more available genecitc basis can make our work closer to real genentics. Also, exploring the latent code relationships of the parent-child datasets might lead to some new genetic evidence.

{\small
\bibliographystyle{ieee_fullname}
\bibliography{egbib}
}

\end{document}